\documentclass[10pt,twocolumn,letterpaper]{article}

\usepackage{cvpr}              % To produce the CAMERA-READY version
\usepackage[greek, english]{babel}
\usepackage{multirow}
\definecolor{cvprblue}{rgb}{0.21,0.49,0.74}
\usepackage[pagebackref,breaklinks,colorlinks,allcolors=cvprblue]{hyperref}

\def\paperID{2594} % *** Enter the Paper ID here
\def\confName{CVPR}
\def\confYear{2025}

\title{T2ICount: Enhancing Cross-modal Understanding for Zero-Shot Counting}

\author{
    Yifei Qian$^1$\thanks{Equal contribution.},
    Zhongliang Guo$^2$\footnotemark[1],
    Bowen Deng$^1$,
    Chun Tong Lei$^3$,
    Shuai Zhao$^4$,
    Chun Pong Lau$^3$,\and
    Xiaopeng Hong$^5$,
    Michael P. Pound$^{1}$\thanks{Corresponding Author.}\and
    {\normalsize$^1$University of Nottingham} \quad
    {\normalsize$^2$University of St Andrews} \quad
    {\normalsize$^3$City University of Hong Kong}\\
    {\normalsize $^4$Nanyang Technology University} \quad
    {\normalsize$^5$Harbin Institute of Technology} \\
    {\small\{yifei.qian, bowen.deng, michael.pound\}@nottingham.ac.uk,
    zg34@st-andrews.ac.uk,}
    {\small\{ctlei2, cplau27\}@cityu.edu.hk,}\\
    {\small shuai.zhao@ntu.edu.sg,
    hongxiaopeng@hit.edu.cn}}
\begin{document}
\maketitle
 \begin{abstract}
Zero-shot object counting aims to count instances of arbitrary object categories specified by text descriptions. Existing methods typically rely on vision-language models like CLIP, but often exhibit limited sensitivity to text prompts. We present T2ICount, a diffusion-based framework that leverages rich prior knowledge and fine-grained visual understanding from pretrained diffusion models. While one-step denoising ensures efficiency, it leads to weakened text sensitivity. To address this challenge, we propose a Hierarchical Semantic Correction Module that progressively refines text-image feature alignment, and a Representational Regional Coherence Loss that provides reliable supervision signals by leveraging the cross-attention maps extracted from the denoising U-Net. Furthermore, we observe that current benchmarks mainly focus on majority objects in images, potentially masking models' text sensitivity. To address this, we contribute a challenging re-annotated subset of FSC147 for better evaluation of text-guided counting ability. Extensive experiments demonstrate that our method achieves superior performance across different benchmarks. Code is available at \url{https://github.com/cha15yq/T2ICount}.

\end{abstract}    
 \section{Introduction}

The task of object counting, estimating the quantity of objects within images, has garnered significant attention due to its broad application across domains~\cite{lempitsky2010learning,sindagi2018survey,qian2023counting}.
Conventional object counting approaches have mainly focused on class-specific counting~\cite{qian2024perspective,zhang2017fcn,tyagi2023degpr, qian2024semi}, requiring extensive annotation and retraining procedures when adapting to novel object categories, reducing general applicability.

\begin{figure}
    \centering
    \includegraphics[width=1\linewidth]{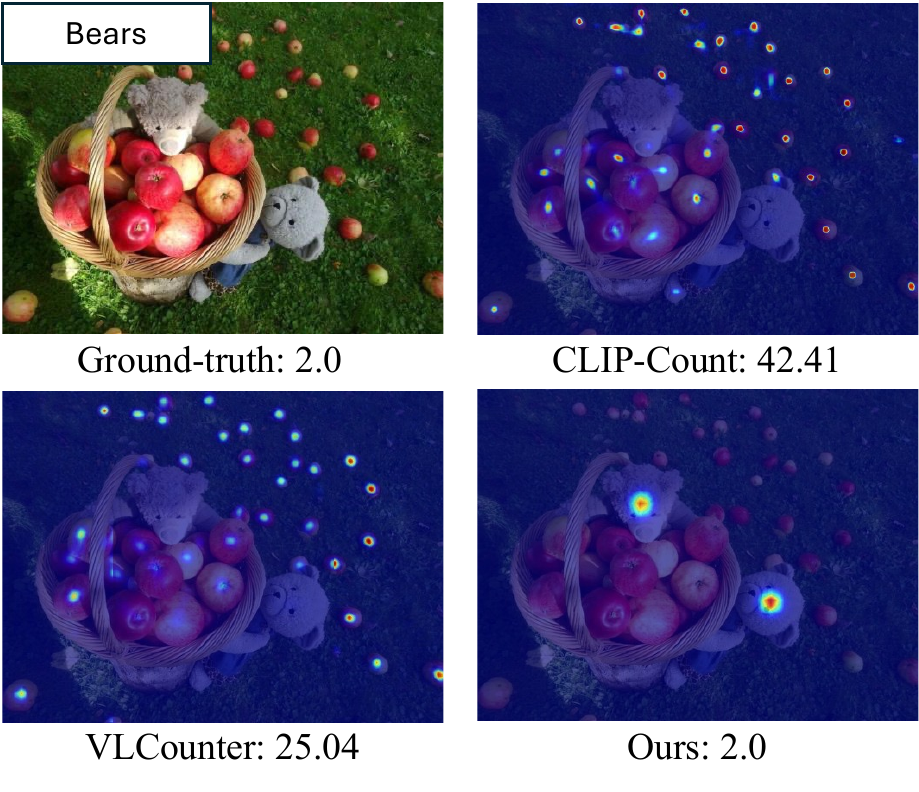}
    \caption{Visualizations of density maps predicted by official pretrained models of two recently proposed text-guided zero-shot object counting methods, CLIP-Count~\cite{jiang2023clip} and VLCounter~\cite{kang2024vlcounter}, which demonstrate poor text sensitivity compared to the proposed T2ICount. }
    \label{fig:problem}
\end{figure}

In contrast, class-agnostic counting encompasses three methodological categories:
\textbf{(a)} \textit{few-shot counting}, which requires a small number of annotated examples per target class;
\textbf{(b)} \textit{reference-less counting}, which estimates quantities based on general object patterns;
\textbf{(c)} \textit{zero-shot counting}, which leverages pre-trained model knowledge to adapt to unseen categories.
The first two paradigms have notable limitations; few-shot methods still require some annotation for new categories~\cite{m_Ranjan-etal-CVPR21, zhizhong2024point, min2022bmnet}, while reference-less approaches lack the ability to focus on specific object categories~\cite{ranjan2022exemplar, hobley2022-LTCA}. Zero-shot methods present a promising direction, enabling counting of specific but previously unseen categories without requiring additional annotation~\cite{jiang2023clip, kang2024vlcounter}.

 Recent approaches to zero-shot counting have relied on pre-trained vision-language models, particularly CLIP~\cite{radford2021learning}, to bridge the semantic alignment gap between visual and textual conditions. These methods focus on fine-tuning CLIP's image encoder to learn counting-specific inductive biases~\cite{jiang2023clip, Amini-Naieni_2023_BMVC, kang2024vlcounter}. As illustrated in Figure~\ref{fig:problem}, however, we observe that these models consistently fail to count text-indicated categories when they differ from the majority class --- \textbf{they remain insensitive to text}. This limitation stems from an inherent feature of the CLIP image encoder: it operates primarily at a global semantic level, naturally biasing the model toward attending to majority object classes described in text prompts rather than minor classes seen within local pixel level information. Counting is an inherently pixel-level task, requiring attention at a local rather than a global level. 

By coincidence, this bias from CLIP aligns with an unintended bias in the commonly used benchmarks such as FSC-147~\cite{min2022bmnet}, in which single annotated labels exist for each image, and the labels overwhelmingly correspond to the numerically dominant object class. The convergence of these two independent factors -- the global semantic bias of CLIP-based models and dataset annotation bias -- creates an illusion of high performance that masks fundamental limitations. The unsolved challenge is therefore to design counting models that are more sensitive to text prompting. A related issue is then how we can mitigate annotation bias to provide a more equitable evaluation of conditional object counting models.

Zero-shot object counting inherently requires deep understanding of text-guided local semantics. A promising solution is to leverage text-to-image diffusion models~\cite{Rombach2022SD, li2023gligen, zhang2023adding, guo2024grey}, which have demonstrated remarkable capability in pixel-level tasks. These models are also trained using CLIP-like text guidance with large-scale pre-training, making them an ideal foundation for zero-shot tasks~\cite{schuhmann2022laionb} with rich prior knowledge that naturally generalizes to diverse, unseen categories in the open world. However, the computationally intensive multi-step denoising process of diffusion models poses significant computational overhead for real-world counting applications.

%Zero-shot object counting inherently requires deep understanding of text-guided local semantics. Hence, we propose to address this problem by re-focusing on local features, by using Diffusion Probabilistic Models (DPMs)~\cite{ho2020denoising}, which exhibit strong performance on pixel level tasks. These models are also trained using CLIP-like text guidance with large-scale pre-training, making them an ideal foundation for zero-shot approach. Our method leverage a text-to-image diffusion model~\cite{Rombach2022SD, li2023gligen, zhang2023adding} pre-trained on LAION-5B~\cite{schuhmann2022laionb}, whose rich prior knowledge naturally generalizes to diverse, unseen categories in the open world. 

In this work, we propose a framework that leverages single-step features from diffusion models to achieve zero-shot counting. Though computationally efficient, this design sacrifices the text awareness that diffusion models typically build through multiple denoising steps~\cite{williams2023unified}, resulting in limited text sensitivity.
%The diffusion model progressively builds up text awareness while preserving fine-grained structural details through its iterative denoising process~\cite{williams2023unified}. %However, the limitation of this approach is that using many iterations is computationally costly, and we argue inappropriate in an application such as object counting. 
%We therefore utilise only a single denoising step, and 
To overcome this limitation, we propose a Hierarchical Semantic Correction Module (HSCM) to compensate for the weakened text-image interaction. The HSCM progressively rectifies the semantic-visual discrepancy through multi-scale feature rectification. We complement this with a novel Representational Regional Coherence Loss ($\mathcal{L}_{\text{RRC}}$) that enhances cross-modal alignment. $\mathcal{L}_{\text{RRC}}$ leverages cross attention maps from the diffusion model to delineate the general foreground regions, thereby solving a key challenge where \textbf{only point-level annotations are available for supervision}: while positive samples can be determined through density thresholds, identifying reliable negative regions is difficult without instance-level annotations. By capturing general object shapes, $\mathcal{L}_{\text{RRC}}$ enables better positive-negative sample selection for more precise feature learning.

To effectively evaluate zero-shot counting performance, we curate FSC-147-S, a specialized subset of FSC-147~\cite{min2022bmnet} designed to provide a more rigorous evaluation protocol for text-guided zero-shot counting. This subset specifically targets scenarios where the text-indicated category differs from the majority class, enabling a more authentic assessment of models' ability to perform category-specific counting beyond the dominant object bias present in existing benchmarks.

% Furthermore, we find that the current benchmark, FSC-147~\cite{min2022bmnet}, is inadequate for assessing a model's counting ability under text guidance, as most images contain only a single object type. This limitation makes it difficult to determine whether a model is truly leveraging textual instructions or simply defaulting to counting the most prominent object~\cite{Amini-Naieni_2023_BMVC}. To solve this, we propose a new evaluation protocol, named FSC-147-S to assess counting performance under text guidance.

In summary, we make the following contributions:
\begin{itemize}
    \item We propose a novel zero-shot object counting framework, T2ICount, leveraging the rich prior knowledge embedded in text-to-image diffusion models.
    \item We identify and address the text insensitivity challenge within zero-shot counting through two key innovations: HSCM for adaptive cross-modal reasoning, and $\mathcal{L}_{\text{RRC}}$ for enhanced visual-language alignment.
    \item We introduce FSC-147-S, a new evaluation protocol that enables contra-bias assessment of conditional counting ability. Our method delivers strong performance on FSC-147 against competing methods, and superior performance on the harder subset of FSC-147-S.
\end{itemize}

% \begin{itemize}
% \item We propose DiffCount, a novel text-guided zero-shot object counting framework which explores to leverage the knowledge embedded in text-to-image diffusion models.
% \item We identify the issue of text insensitivity in the single-step denoising process and tailor a Semantic-Powered Explicit Cross-scale TUning Module, along with a Representational Regional Coherence Loss to strengthen the visual-language association.
% \item We propose an evaluation protocol to assess the model's counting ability under text guidance. Extensive experiments show the effectiveness of DiffCount.
% \end{itemize}

 \section{Related Work}

\begin{figure*}[t]
    \centering
    \includegraphics[width=0.95\linewidth]{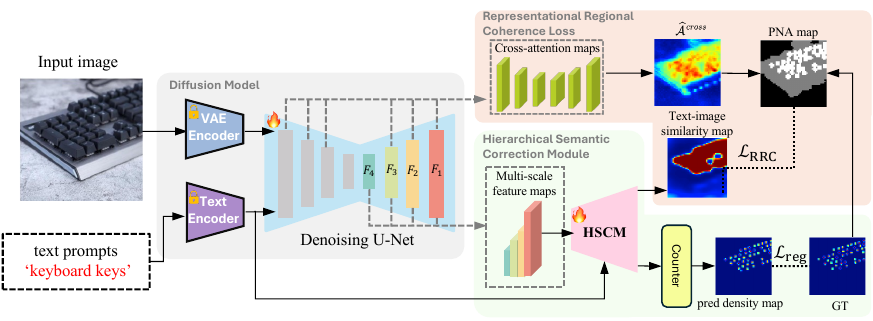}
    \caption{\textbf{Overview of the proposed T2ICount.} Our method is based on single denoising step. An input image and text prompts specifying the category to be counted are fed into the pre-trained text-to-image diffusion model. Feature maps extracted from the decoder of the U-Net are passed through the Hierarchical Semantic Correction Module to enhance textual awareness, producing the final features used to estimate the density map. Text-image similarity maps are generated at intermediate stages and are supervised by the  Representational Regional Coherence Loss. The ground-truth density map and the fused cross attention maps ($\widehat{\mathcal{A}}^{cross} $) are used to generate the positive-negative-ambiguous (PNA) map, providing supervision signals for this loss. In the training process, the VAE encoder and the text encoder are frozen while the U-Net and HSCM are being trained.}
    \label{fig:Overall-framework}
\end{figure*}

\subsection{Few-shot Object Counting}
Few-shot object counting aims to train a generalised counting model that can estimate the number of objects of an arbitrary class given a few visual exemplars during inference. This problem is formulated as a matching task in the pioneering work GMN~\cite{lu2019class}, which uses a two-stream architecture to explore the similarity between image and exemplar features for counting. The subsequent work~\cite{m_Ranjan-etal-CVPR21}, FamNet adopts a single-stream architecture with ROI pooling to extract exemplar features. In addition, to address the lack of suitable datasets for this task, a new multi-class dataset, FSC-147 is now commonly used to evaluate object counting tasks. Later research has focused on either improving the quality of feature representations through advanced backbone architectures~\cite{liu2022countr, Dukic_2023_ICCV}, or optimizing the matching mechanism by enhancing the image-exemplar feature similarity map~\cite{min2022bmnet, you2023few, lin2022scale}. While significant progress has been made in few-shot object counting, the task remains reliant on manually-provided exemplars. These can be costly to obtain, and may introduce bias by not fully capturing the diversity and variability of the target objects.

% \subsection{Reference-less Object Counting}
%  Reference-less object counting has been proposed as an alternative which counts instances of the the dominant class in an image without any exemplars. In pioneering work, RepRPN-Counter~\cite{ranjan2022exemplar} proposes a module that selects the most frequently repeating objects from the image as a substitute for exemplars. RCC~\cite{hobley2022-LTCA} offers a solution using a pre-trained ViT to directly regress object counts with weak image-level supervision. Despite this progress, a key limitation of reference-less counting is the inability to specify which objects to count, particularly in complex scenes containing multiple object types.

\subsection{Zero-shot Object Counting}
Zero-shot object counting addresses this issue by incorporating external information, such as text descriptions, allowing the model to count specific object categories without exemplars. Xu \textit{et al.}~\cite{Xu_2023_CVPR} propose the first method for this task, building on a few-shot counting framework in which a text-conditioned variational autoencoder (VAE) is used to generate visual exemplars specified by the text. Later, three concurrent workshave leveraged the association between text and image embeddings learned by the pre-trained language-vision foundation model, CLIP~\cite{radford2021learning}, for text-guided zero-shot counting. These are CLIP-Count~\cite{jiang2023clip}, VLCounter~\cite{kang2024vlcounter}, and CountX~\cite{Amini-Naieni_2023_BMVC}, respectively. All three methods demonstrate similar counting performance but remain far behind state-of-the-art visual exemplar-based frameworks. More recently, new methods have been proposed that attempt to adapt different vision-language models for this task. PseCo~\cite{zhizhong2024point} introduces a framework that leverages the Segment Anything model~\cite{kirillov2023segment} for proposal generation and uses CLIP for classification based on text specification. VA-Count~\cite{zhu2024zero} leverages Grounding DINO~\cite{liu2023grounding} for initial text-specified detection, then proposes a mechanism to select good visual exemplars from its predictions for match-based counting. Despite rapid progress, a key challenge remains in assessing how well the model is counting the objects specified by the text, largely due to the characteristics of the FSC-147 dataset, in which all images are annotated with only a single object class, and in most cases this is the majority class. To overcome this, we re-annotated a portion of the FSC-147 dataset specifically designed to evaluate the model's behavior in text-guided counting.

%\subsection{Diffusion models}
%Diffusion Denoising Probabilistic Models~\cite{ho2020denoising} and their variant, Latent Diffusion Models~\cite{Rombach2022SD}, have recently gained significant attention in generative tasks for their ability to generate high-quality images by gradually denoising noisy data. These models are powerful due to the U-Net structure which captures the inductive biases of image-like data, and the cross-attention mechanism that bridges various conditioning inputs~\cite{zhang2023adding, wang2024diffx}, such as text and semantic maps with visual features. Unlike CLIP features, which are trained to capture global semantics, diffusion models are specifically tailored for pixel-level tasks, making their features inherently more suitable for object counting. However, due to the nature of their training objective, diffusion model features may not be as sensitive to text input. Therefore, in this work, we explore how to adapt the prior knowledge in diffusion models for text-guided zero-shot object counting, while also attempt to address their limitations in this task.

 \section{Methodology}
\begin{figure*}[t]
    \centering
    \includegraphics[width=\linewidth]{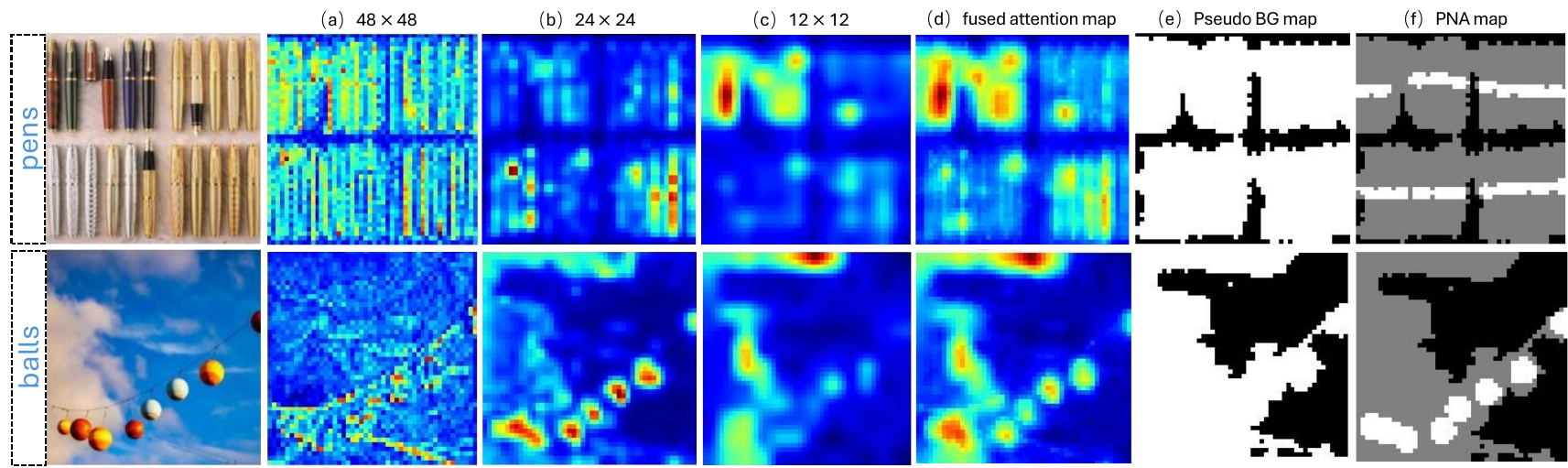}
    \caption{Visualization of the issue of text sensitivity and key maps in supervision signal generation of $\mathcal{L}_{\text{RRC}}$.  \textbf{(a-c)} Cross-attention maps from different layers of pre-trained Stable Diffusion v1.5~\cite{Rombach2022SD}, demonstrating weak text-image sensitivity in single-step denoising; \textbf{(d-f)} Key intermediate maps for constructing supervision signals: \textbf{(d)} fused cross-attention map, \textbf{(e)} derived pseudo-background map (white: foreground, black: background), and \textbf{(f)} positive-negative-ambiguous map (white: positive, black: negative, gray: ambiguous regions)}
    \label{fig:text_insensi}
\end{figure*}

The goal of text-guided zero-shot object counting is to estimate a density map $d =f(x, c)$ that describes the count of arbitrary types of objects in an image $x$, specified by the input text $c$, where the object types are not restricted to those in the training set. Specifically, we aim to learn a mapping function $f: \mathcal{X} \times \mathcal{T} \rightarrow \mathcal{D}$, which maps from the image space $\mathcal{X}$ and text space $\mathcal{T}$ to the density map space $\mathcal{D}$.

Here we outline T2ICount, a novel framework leveraging pretrained diffusion models for zero-shot object counting, as illustrated in Fig~\ref{fig:Overall-framework}. By extracting features from a single denoising step rather than the full diffusion process~\cite{ranasinghe2024crowddiff}, our framework achieves practical efficiency for real-world applications. However, we observe that this efficiency comes at the cost of weak sensitivity between image and text features. We first analyze this issue in Sec.~\ref{sec:text:insens}, then introduce the Hierarchical Semantic Correction Module (Sec.~\ref{sec:HSCM}) and the Representational Regional Coherence Loss (Sec~\ref{sec:RRCL}) to address this challenge.

\subsection{Text Insensitivity in Single-Step Denoising}\label{sec:text:insens}
Diffusion models have emerged as a powerful family of generative models~\cite{zhang2025, csiimd} that learn complex data distributions through a gradual denoising process. In our framework, we leverage Stable Diffusion~\cite{Rombach2022SD}, which performs diffusion in a compact latent space in which images and text are encoded via a VAE encoder and CLIP encoder respectively. Formally, given a latent variable $z_t$ corrupted from the compressed image representation $z_0$ through a forward diffusion process:
\begin{equation}
\begin{split}
z_t = \sqrt{\bar{\alpha}_t}z_0 + \sqrt{1-\bar{\alpha}_t}\epsilon,\\
\bar{\alpha}_{t}=\prod_{i=1}^{t}\alpha_{i}, \epsilon\sim\mathcal{N}(\mathbf{0},\mathbf{I}),    
\end{split}    
\end{equation}
where $\bar{\alpha}_{i}$ controls the amount of noise and $\epsilon$ is random noise drawn from a standard normal distribution. The denoising U-Net iteratively predicts and removes the noise $\epsilon_\theta$ conditioned on encoded text features. This iterative denoising process enables the progressive alignment between text and visual representations. 

For computational efficiency in object counting, our method utilizes single-step diffusion features. However, this inherently constrains the model's capacity to establish robust text-vision correspondence. This limitation is particularly evident when leveraging a small denoising step ($t=1$), selected to maintain proximity to the original representation. As demonstrated in Fig.~\ref{fig:text_insensi}: (a-c),  the resulting cross-attention maps exhibit substantial semantic misalignment, in which regions irrelevant to the text prompt are highlighted, and with inconsistent attention on semantically relevant objects. The degradation is more severe in low-level feature maps, which display heightened noise characteristics.

%In diffusion models, text awareness builds up over multiple denoising steps, where each step gradually aligns the text and image features for text-image coherence. However, in our task, we prioritize the computational efficiency and thus utilize only a single denosing step. This choice significantly reduces the computational load but also limits the model's opportunity to fully incorporate textual guidance, leading to weaker text awareness. This limitation is especially pronounced when using the late denoising step, which we select to keep the image as close to the original as possible. As shown in Fig.~\ref{fig:text_insensi}: (a-c), all of these cross-attention maps highlight areas irrelevant to the text and fail to assign consistent values to the relevant objects. Furthermore, the low-level cross-attention maps appear to be much noisier than the high-level ones. This misalignment becomes even more pronounced when the prompt is entirely unrelated to the image, as the cross-attention maps still highlight general foreground regions rather than responding appropriately to the text. 

To address these limitations, we naturally decompose the objective mapping function $f$ into a composition of mappings: $f = \mathcal{G}(\mathcal{H}(x, \mathcal{Q}(c)), \mathcal{Q}(c))$. Here, $\mathcal{Q}:\mathcal{T}\rightarrow\mathcal{C}$ represents the CLIP text encoder that maps input text $c$ from the text space $\mathcal{T}$ to the text feature space $\mathcal{C}$, while $\mathcal{H}:\mathcal{X}\times\mathcal{C}\rightarrow\mathcal{F}$ is implemented by $\epsilon_\theta$, which integrates the image $x$ and the text features $c'$ to produce feature maps $\mathcal{F}$. Finally, $\mathcal{G}:\mathcal{F}\times\mathcal{C}\rightarrow\mathcal{D}$ maps $\mathcal{F}$ and $\mathcal{C}$ to the density map space $\mathcal{D}$. In the following sections, we will focus on the design of $\mathcal{G}$, which we implement as a Hierarchical Semantic Correction Module guided using a Representational Regional Coherence Loss.

\subsection{Hierarchical Semantic Correction Module}\label{sec:HSCM}
The HSCM progressively refines text-image feature alignment through a multi-stage process. Given four extracted hierarchical multi-scale feature maps $F_i \in \mathbb{R}^{\frac{H}{2^{i-1}} \times \frac{W}{2^{i-1}}}$ ($i\in \{1,2,3,4\}$) of decreasing resolution and text features $c'$, where $H$ and $W$ denote the spatial dimensions of the latent vector $z_t$, the module operates in three stages. At each stage, feature maps from adjacent levels are fused as:
\begin{equation}
    F^{'}_i =  \texttt{Conv}\big(\texttt{Concat}\big(\texttt{Up}\big( V_{i+1} \big), F_i\big)\big),
\end{equation}
where $F^{'}_i$ is the fused feature map, $V_i$ denotes features from the previous stage (with $V_4 = F_4$), $\texttt{Conv}$, $\texttt{Concat}$, and $\texttt{Up}$ represent convolution, channel-wise concatenation, and upsampling operations, respectively.

The success of our multi-stage refinement relies on two modules that address different aspects of semantic alignment: the Semantic Enhancement Module (SEM) and the Semantic Correction Module (SCM).

\noindent \textbf{Semantic Enhancement Module (SEM)}: The SEM facilitates bidirectional cross-modal interaction through text-to-image and image-to-text attention mechanisms. We then compute a text-image similarity map between the enhanced feature maps $V_i$ and text representation $c'$ to measure cross-modal alignment:
\begin{equation}
    S_i= \frac{V_i \cdot c'}{\|V_i\| \|c'\|}.
\end{equation}
Supervised by $\mathcal{L}_{\text{RRC}}$, the SEM learns to generate text-image similarity maps $S_i$ that capture class-specific object regions like segmentation masks. This mask serves as attention guidance in SCM to highlight semantically relevant regions.

\noindent \textbf{Semantic Correction Module (SCM)}: The SCM rectifies feature representations by incorporating similarity-guided features from the previous stage:
\begin{equation}
F^{'}_i + \texttt{Up}\big( V_{i+1} \odot S_{i+1} \big) \rightarrow F^{'}_i,
\end{equation}
where $\odot$ denotes element-wise multiplication. This module redirects the model's attention to these text-relevant regions, facilitating density map learning for counting.

The refinement process varies across different stages to progressively enhance text-image understanding. Specifically:
In the starting stage ($i=3$), we apply the SEM to $F'_3$ to obtain $V_3$. For the intermediate stage ($i=2$), the fused features $F'_2$ first undergo SCM correction followed by SEM, producing $V_2$. In the final stage ($i=1$), we apply the SCM to $F'_1$ to prepare the final features $V_1$ for counter. Through this cascaded design, each stage corrects and refines features from previous stages, ensuring progressively enhanced text-image alignment for counting.

\subsection{Representational Regional Coherence Loss}\label{sec:RRCL}
The supervision on each intermediate text-image similarity map $S_i$ is crucial for refining regional text-image coherence. However, the lack of instance-level annotations in counting datasets poses a significant challenge in identifying positive and negative samples from point-level annotations. Traditional methods~\cite{jiang2023clip, kang2024vlcounter} typically rely on a simple density thresholding strategy: regions with density values above a threshold are considered positive, while others are treated as background. However, this naive approach inevitably misclassifies many foreground regions. This leads to inconsistent semantic supervision, hindering the model from learning accurate text-image alignments. 

To address this issue, we derive robust supervision signals by leveraging cross-attention maps from the diffusion model. Interestingly, as shown in Fig.~\ref{fig:text_insensi}: (a-c), while single-step attention maps show weak sensitivity to specific object categories mentioned in the text, they effectively capture the overall foreground regions in the image. Based on this observation, we leverage these attention maps to identify background pixels. 

Specifically, we first extract cross-attention maps $\mathcal{A}_i^{cross}$ at different spatial scales from $\epsilon_\theta$ and unify their resolutions through upsampling. These maps are then fused using appropriate weights, $w_i$, to obtain a fused attention map ${\mathcal{A}}^{cross}$, which can be expressed as:

\begin{equation}
    {\mathcal{A}}^{cross} = \sum_{i}w_i\cdot\text{norm}(\mathcal{A}_i^{cross}) \in \mathbb{R}^{H\times W},
\end{equation}

where $\text{norm}$ indicates min-max normalization. The obtained ${\mathcal{A}}^{cross}$, together with the ground-truth density map $D^{\text{gt}}$, is used to generate a Positive-Negative-Ambiguous (PNA) map that provides supervision signals, where values 1, 0, and -1 indicate positive, negative, and ambiguous regions respectively. Formally, for each position $(j,k)$ in the PNA map, the value $p_{jk}$ is determined as:

\begin{equation} 
p_{jk} = \left\{ \begin{aligned} 1, & \quad \text{if} \quad D^{\text{gt}}_{jk} \geq \tau, \\ 0, & \quad \text{else if} \quad {\mathcal{A}}^{\text{cross}}_{jk} \leq \theta, \\ -1, & \quad \text{otherwise}.  \end{aligned} \right. 
\label{eq:pjk}
\end{equation}

Here, $\tau$ and $\theta$ are thresholds. We visualize some intermediate results in our supervision signal generation in Fig~\ref{fig:text_insensi}: (d) ${\mathcal{A}}^{\text{cross}}$, (e) pseudo-background maps obtained by directly binarizing the ${\mathcal{A}}^{\text{cross}}$ using the second condition in Eq~\ref{eq:pjk}, and (f) PNA maps. Note that these ambiguous regions (shown in gray) are commonly treated as negative regions in traditional methods. 

With the PNA map, the $\mathcal{L}_{\text{RRC}}$ is defined as follows:
\begin{equation}
    \mathcal{L}_{\text{RRC}} = \lambda\mathcal{L}_{\text{pos}} + \mathcal{L}_{\text{neg}}
\end{equation}
where
\begin{equation}
    \mathcal{L}_{\text{pos}} = \sum_{jk}1 - S_{jk} \quad \text{if} \ \ p_{jk}=1,
\end{equation}
and
\begin{equation}
    \mathcal{L}_{\text{neg}} = \sum_{jk} \texttt{max}(0, S_{jk}) \quad \text{if} \ \ p_{jk}=0.
\end{equation}
Here, $\lambda$ is a balancing factor. There is no explicit restrictions imposed on ambiguous regions. The overall loss function for training T2ICount is given below:
\begin{equation}
    \mathcal{L} = \mathcal{L}_{\text{reg}} + \gamma\mathcal{L}_{\text{RRC}}
\end{equation}
where $\mathcal{L}_{\text{reg}}$ refers to the regression loss which we directly adopt the same with that used in CUT~\cite{qian2022segmentation} and $\gamma$ is a balancing factor. 

 \section{Experiments}
\begin{table*}[t]
  \centering
  \caption{Performance comparison of T2ICount with other state-of-the-art models on FSC-147 dataset. The best performance for each scheme is highlighted in \textbf{bold}, and the second-best performance for the zero-shot setting is \underline{underlined}.}
    \begin{tabular}{l|c|c|c|cc|cc}
    \toprule
    \multirow{2}{*}{Scheme}&\multirow{2}{*}{Method} & \multirow{2}{*}{Venue} & \multirow{2}{*}{Shot} & \multicolumn{2}{c|}{Val Set} & \multicolumn{2}{c}{Test Set}\\\cline{5-8} 
    &&&&MAE&RMSE&MAE&RMSE\\
    \hline
    \multirow{8}{*}{Few-shot}& FamNet~\cite{m_Ranjan-etal-CVPR21}&CVPR'21 &3&  24.32 & 70.94 &22.56& 101.54\\
    %&CFOCNet~\cite{Yang2021WACV}&WACV'21&3&21.19 &61.41 &22.10 &112.71\\
    %&CounTR~\cite{liu2022countr}& BMVC'22&3& 13.13& 49.83& 11.95& 91.23\\
    &BMNet~\cite{min2022bmnet}&CVPR'22&3& 15.74 &58.53 &14.62& 91.83\\
    &LOCA~\cite{Dukic_2023_ICCV}&ICCV'23&3&\textbf{10.24} &\textbf{32.56}& \textbf{10.97}& \textbf{56.97}\\
    &SAM~\cite{shi2024training}&WACV'24&3&-& - &19.95& 132.16\\
    &PseCo~\cite{zhizhong2024point}&CVPR'24&3&15.31 &68.34& 13.05& 112.86\\ \cline{2-8}
    &GMN~\cite{lu2019class}&ACCV'19&1&29.66& 89.81& 26.52& 124.57\\
    &FamNet~\cite{m_Ranjan-etal-CVPR21}&CVPR'21&1&24.32 &70.94 &22.56& 101.54\\
    &BMNet~\cite{min2022bmnet}&CVPR'22&1& 19.06 &67.95 &16.71& 103.31\\
    \hline
    \multirow{3}{*}{Reference-less} &FamNet~\cite{m_Ranjan-etal-CVPR21}&CVPR'21&0&32.15 &98.75& 32.27& 131.46\\
    %&RepRPN-C~\cite{ranjan2022exemplar}&ACCV'22&0& 29.24 &98.11& 26.66& 129.11\\
    %&CounTR~\cite{liu2022countr}& BMVC'22&0&18.07& 71.84& \textbf{14.71}& 106.87\\
    &LOCA ~\cite{Dukic_2023_ICCV}&ICCV'23&0& \textbf{17.43}& \textbf{54.96}& 16.22& \textbf{103.96}\\
    &RCC~\cite{hobley2022-LTCA}& CVPR'23&0& 17.49 &58.81& 17.12& 104.53\\
    \hline
    \multirow{8}{*}{Zero-shot}& 	
Patch-selection~\cite{Xu_2023_CVPR}&CVPR'23&0&26.93 &88.63& 22.09& 115.17\\
    &CLIP-Count~\cite{jiang2023clip} & ACM MM’23& 0& 18.79& 61.18& 17.78& 106.62\\
    &CounTX~\cite{Amini-Naieni_2023_BMVC} & BMVC'23&0&17.10 &65.61& 15.88& 106.29\\
    &VLCounter~\cite{kang2024vlcounter}&AAAI'24 &0&18.06& 65.13& 17.05& 106.16\\
    &PseCo~\cite{zhizhong2024point}&CVPR'24& 0 &23.90 &100.33& 16.58& 129.77 \\
    &DAVE~\cite{pelhan2024dave}&CVPR'24&0&15.48 &\textbf{52.57}& 14.90 & \underline{103.42}\\ 
    &VA-Count~\cite{zhu2024zero}&ECCV'24&0&17.87& 73.22& 17.88& 129.31\\
    &GeCo~\cite{pelhan2024novelunifiedarchitecturelowshot} &NeurIPS'24& 0& \underline{14.81} & 64.95 & \underline{13.30} & 108.72\\
    &T2ICount (Ours)&-&0&\textbf{13.78} &\underline{58.78}& \textbf{11.76}&\textbf{97.86}\\
    \bottomrule
  \end{tabular}
  \label{tab:result}
\end{table*}
\subsection{Dataset and Evaluation Metrics}

We evaluate the proposed T2ICount on FSC-147~\cite{min2022bmnet} and CARPK~\cite{Meng2017Car}. FSC-147 dataset contains 6135 paired images of 147 object classes. which is designed for class-agnostic counting. It is split into 3,659 training, 1,286 validation, and 1,190 test images, with non-overlapping classes across splits, making it well-suited for the zero-shot object counting task. The class names are directly used as the text input $c$. We use the CARPK dataset to evaluate the generalizability of T2ICount, it contains 1,448 images of car parks captured by drones. %ShanghaiTech is a human counting dataset that has two parts, A and B. Part A contains 482 images, with 300 designated for training and 182 for testing, while part B comprises 716 images, with 400 for training and 316 for testing. Notably, part A has a much higher crowd density compared with part B. We use `car' and `human' as the text prompts for CARPK and ShanghaiTech, respectively

\textbf{Evaluation Protocol for Text-Guided Counting Performance}: The FSC-147 dataset primarily features simple scenes, typically containing a single, highly prominent object type per image with relatively plain backgrounds. This simplicity limits its effectiveness in evaluating a model's ability to perform text-guided counting in complex scenarios, particularly given that previous methods~\cite{ranjan2022exemplar} for reference-less counting have demonstrated that models can count the most repetitive objects in an image without relying on visual or text prompts. To address this, we manually extract a subset from FSC-147, named FSC-147-S, that focuses on images that contain at least two object categories. We then supplement this subset with count annotations for a less frequent category, which is typically present in significantly lower quantities than the primary object category. This subset contains 196 images, with the average count for the less frequent class at 4.6, in contrast to the original annotated classes which average around 49.4 instances. This intentionally imbalanced setup challenges the model to rely on the guidance of the prompt, rather than merely identifying and counting the most frequent or repetitive objects. This provides a clearer assessment of text-guided counting capability.

Following previous works on object counting, we evaluate the performance of our method using mean absolute error (MAE) and root mean squared error (RMSE) metrics.

\subsection{Implementation Details}
\textbf{Architecture Detail of Counter:} The Counter module generates density maps by first applying self-attention operations on the input feature map. The attended features are then passed through three convolutional layers to produce the estimation.

\noindent\textbf{Training:} We train the proposed T2ICount model on the training set of FSC-147. Our model is initialized from the pre-trained Stable Diffusion v1.5~\cite{Rombach2022SD}, with the VAE decoder removed. We fix the weights of the VAE encoder and the CLIP text encoder, while fine-tuning the weights of the U-Net to learn the counting-specific task. The base learning rate is set to $5\times10^{-5}$. To better preserve the pre-trained knowledge, the learning rate for the U-Net is reduced to $1/10$ the base learning rate. We train the model for 400 epochs using the AdamW optimizer with a weight decay of $1\times10^{-4}$ and a batch size of 16, on a single NVIDIA RTX A6000 GPU. We apply the same data augmentations as used in CounTX~\cite{Amini-Naieni_2023_BMVC} except for the gaussian blur. We also implement random rescaling with a factor within [1, 2]. Regarding the hyperparameters in our framework, we set $\lambda$ and $\gamma$ as 2 and 0.01, respectively. To generate $\mathcal{A}^{cross}$,  we empirically set $w_i$ for the cross-attention maps at sizes of 12$\times$12, 24$\times$24, and 48$\times$48 as [0.6, 0.3, 0.1], respectively.
\begin{figure*}[t]
    \centering
    \includegraphics[width=\linewidth]{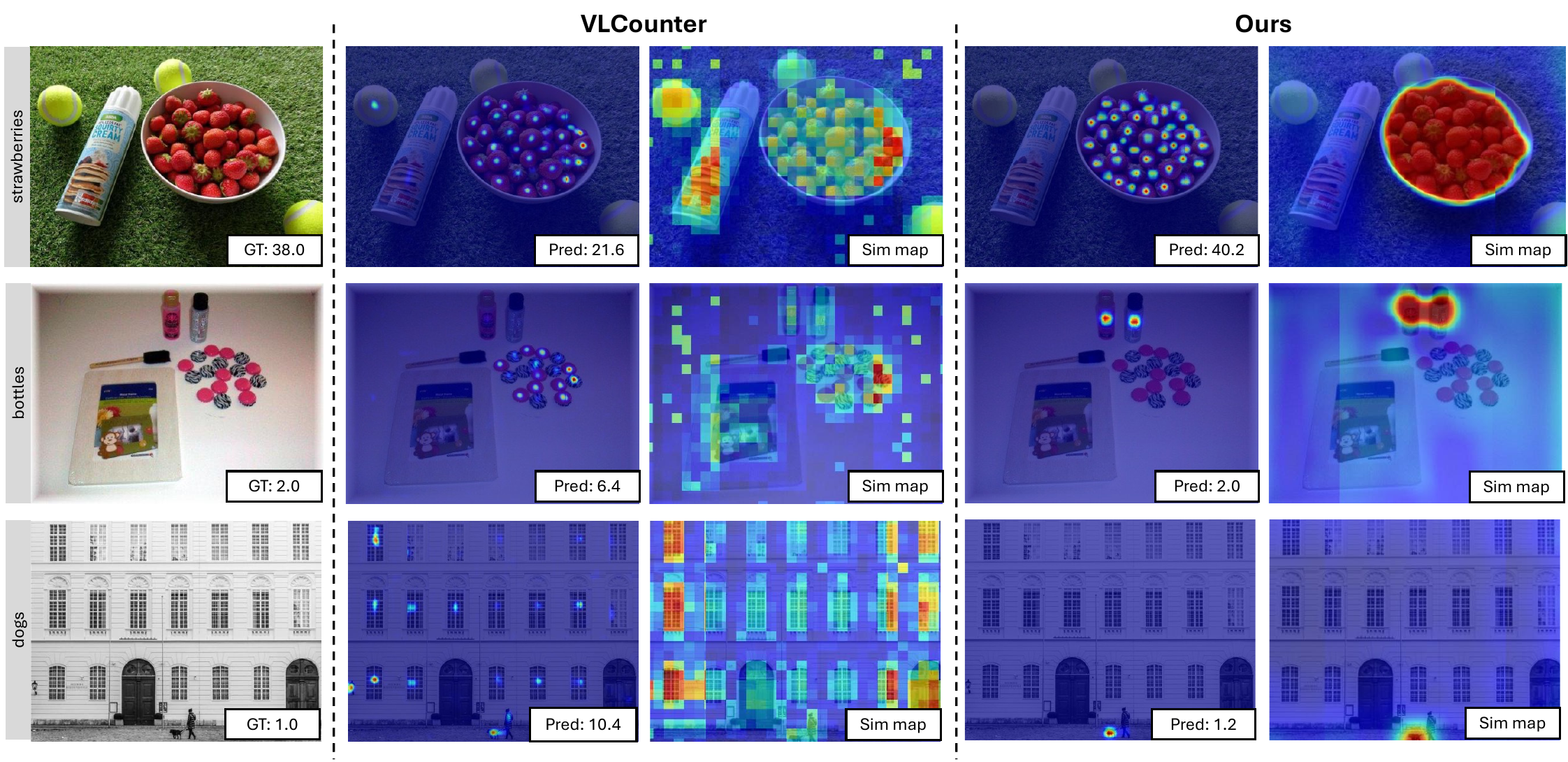}
    \caption{Qualitative comparison of T2ICount with VLCounter~\cite{kang2024vlcounter}. With our proposed $\mathcal{L}_{\text{RRC}}$, our text-image similarity map exhibits reduced noise and more precise object delineation, which results a more accurate density estimation. }
    \label{fig:vis-result2}
\end{figure*}

\noindent\textbf{Inference:} We employ a sliding window of size 384$\times$384 with a stride of 384 to scan over the entire image. For overlapping regions, the density map values are computed by averaging.

\subsection{Comparison with the State-of-the-Arts}
We benchmark the performance of our method against various few-shot, reference-less, and zero-shot object counting approaches.

\textbf{Quantitative Result on FSC-147:} The result is reported in Table~\ref{tab:result}. We demonstrate that even when trained solely with text prompts, our model achieves competitive performance compared to few-shot learning and reference-less counting methods. When compared to other text-specified zero-shot object counting methods, T2ICount attains state-of-the-art performance, achieving the lowest MAE and RMSE on the test set, along with the lowest MAE and second-lowest RMSE on the validation set. Moreover, our diffusion-based approach surpasses the previous best CLIP-based counting method, CounTX, with further reductions of 25.9\% in MAE and 7.9\% in RMSE on the test set. 

\begin{table}[!htbp]
    \centering
    \caption{Comparison of T2ICount with other state-of-the-art zero-shot object counting models on the FSC-147-S dataset.}
    \begin{tabular}{l|cc}
    \toprule
      Method   & MAE& RMSE \\ \hline
      CLIP-Count~\cite{jiang2023clip}   &  48.42 & 108.04\\
      CounTX~\cite{Amini-Naieni_2023_BMVC} & \underline{31.30} & 98.80 \\
      VLCounter~\cite{kang2024vlcounter}& 35.24&75.46 \\
      PseCo~\cite{zhizhong2024point}&39.01 &\underline{61.34}\\
      DAVE~\cite{pelhan2024dave} &49.32 & 108.47\\
      T2ICount (Ours) & \textbf{4.69}& \textbf{8.06}\\
         \bottomrule
    \end{tabular}
    \label{tab:147-S}
\end{table}

\textbf{Quantitative Result on FSC-147-S:}  We compare T2ICount with three state-of-the-art CLIP-based methods—CLIP-Count~\cite{jiang2023clip}, CountX~\cite{Amini-Naieni_2023_BMVC}, and VLCounter~\cite{kang2024vlcounter}—on the new evaluation protocol, FSC-147-S. As shown in Table~\ref{tab:147-S}, our method achieves the best performance, significantly reducing MAE by 85.1\% and RMSE by 86.9\%, respectively. The strong results suggest that our method adheres closely to the guidance provided by the text prompt, accurately focusing on the specified object for counting.
\begin{table}[!htbp]
    \centering
    \caption{Comparison of T2ICount with other state-of-the-art zero-shot object counting models on the CARPK dataset.}
    % \resizebox{0.65\linewidth}{!}{
    \begin{tabular}{l|cc}
    \toprule
      Method   & MAE& RMSE \\ \hline
      RCC~\cite{hobley2022-LTCA} & 21.38& 26.61 \\
      CLIP-Count~\cite{jiang2023clip}   &  11.96 & 16.61\\
      CounTX~\cite{Amini-Naieni_2023_BMVC} & \textbf{8.13}& \textbf{10.87} \\
      %VLCounter~\cite{kang2024vlcounter}& 6.46& 8.68\\
      Grounding DINO~\cite{liu2023grounding} & 29.72 & 31.60\\
      VA-Count~\cite{zhu2024zero} & 10.63 & \underline{13.20}\\
      T2ICount (Ours) &\underline{8.61} &13.47 \\
         \bottomrule
    \end{tabular}
    \label{tab:carpk}
\end{table}

\textbf{Quantitative Result on CARPK:} We assess the cross-dataset generalizability of our model, trained on FSC-147, by testing it on the CARPK dataset. The results are reported in Table~\ref{tab:carpk}. Our method demonstrates competitive performance, achieving the second lowest MAE, indicating strong adaptability across datasets.

\begin{figure*}[!htbp]
    \centering
    \includegraphics[width=\linewidth]{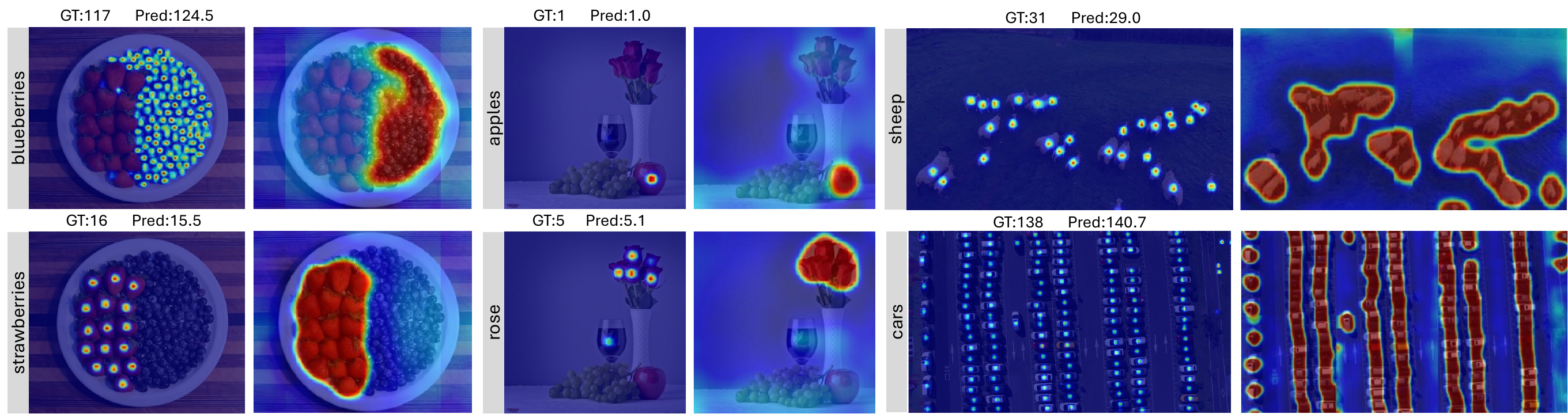}
    \caption{Qualitative results of T2ICount. Each pair shows the predicted density map (left) and the corresponding text-image similarity map (right), where the similarity maps effectively delineate the overall shapes of text-specified objects.}
    \label{fig:vis-result}
\end{figure*}

\subsection{Ablation Study}
We conduct a series of ablation studies on the components of T2ICount. The baseline model is defined as comprising only the stable diffusion model components and the counter structure where the counter directly performs on $F_4$ to make predictions under the supervision of $\mathcal{L}_{\text{reg}}$. In this setup, we apply the $\mathcal{L}_{\text{RRC}}$ on $F_4$ to verify the effectiveness of our designed loss. Finally we integrate the proposed HSCM. The performance is evaluated on both the FSC-147 test set and the introduced FSC-147-S. The results are presented in Table~\ref{tab:ablation}. Firstly, the baseline model's performance on the test set highlights the high quality of features derived from the pre-trained diffusion model. However, its results on FSC-147-S highlight the limitations in effectively aligning the text prompt with the visual representations. This also highlights how the base FSC-147 dataset can mask some inadequacies in the method. Then with the $\mathcal{L}_{\text{RRC}}$ added, we observe substantial improvements on FSC-147-S, achieving reductions of approximately 60.6\% in MAE and 65.63\% in MSE. However the performance on the FSC-147 test set only increases a small amount. Finally, with the HSCM included, the model benefits from enriched feature detail and a more progressive text-image alignment process, resulting in further reductions in MAE and MSE by 19.14\% and 7.86\% on the FSC-147 test set, and by 51.09\% and 63.78\% on FSC-147-S. 

%\noindent\textbf{Effects of the Text-Image Alignment Loss:}

\subsection{Qualitative Results}
Fig~\ref{fig:vis-result2} shows qualitative comparisons between T2ICount and VLCounter~\cite{kang2024vlcounter} through their predicted density maps and text-image similarity maps. Each density map and similarity map is overlaid on top of its corresponding image. Thanks to the guidance of $\mathcal{L}_{\text{RRC}}$, our text-image similarity map achieves high-quality object delineation, effectively capturing the overall shape of target objects rather than fragmenting into task-specific regions that could impair semantic understanding. In contrast, VLCounter's approach of treating low-density regions as negative samples results in poor semantic alignment and substantial noise in similarity maps. We present more visualization results in Fig~\ref{fig:vis-result}. The text-image similarity map captures the holistic semantic understanding of target objects, which guides our model to generate precise density maps for accurate counting predictions. The last example in Fig~\ref{fig:vis-result} shows results on the CARPK dataset, demonstrating T2ICount's generalization capability across different domains. To conclude, our model effectively distinguishes between object classes based on their textual descriptions.

\begin{table}[!htbp]
    \centering
    \caption{Ablation study on the key components of T2ICount}
    % \resizebox{0.87\linewidth}{!}{
    \begin{tabular}{l|cc|cc}\toprule
    \multirow{2}{*}{}& \multicolumn{2}{c|}{Test} & \multicolumn{2}{c}{FSC-147-S}\\\cline{2-5}
    & MAE & RMSE & MAE & RMSE\\\hline
    Baseline~(B)     & 14.66 & 111.62 & 24.34 & 64.74\\
    B + $\mathcal{L}_{\text{RRC}}$ &  14.55 &106.21 & 9.59 & 22.25\\
    B + $\mathcal{L}_{\text{RRC}}$ + HSCM& \textbf{11.76}&\textbf{97.86} & \textbf{4.69} & \textbf{8.06}\\\bottomrule
    \end{tabular}
    \label{tab:ablation}
\end{table}
 \section{Conclusion}

In this paper we have presented T2ICount, a new approach to zero-shot object counting. Our approach directly addresses the challenge of text insensitivity prevalent among text-guided counting models. We design a Hierarchical Semantic Correction Module for progressive feature refinement, and a Representational Regional Coherence Loss for reliable supervision. Extensive experiments show that our method achieves superior performance on current benchmarks. We also reveal the evaluation bias found within existing benchmarks, and contribute a re-annotated subset of FSC147 for more effective assessment of text-guided counting ability. On this harder task, our method out-competes others by a wide margin. Our future work will focus on constructing a more diverse dataset with richer object categories to further advance text-guided counting research.

 \section*{Acknowledgements}

This work was supported by the Biotechnology and Biological Sciences Research Council (grant number BB/Y513908/1). This work was also funded by the National Natural Science Foundation of China (62376070, 62076195).
{
    \small
    \bibliographystyle{ieeenat_fullname}
    \bibliography{main}
}

\end{document}